\documentclass[letterpaper, 10pt, conference]{ieeeconf}
\IEEEoverridecommandlockouts 
\usepackage{amsmath,amssymb,url}
\usepackage{graphicx}
\usepackage{xcolor}
\usepackage{siunitx}
\usepackage[hidelinks]{hyperref}
\usepackage{cleveref}
\usepackage{booktabs}

\usepackage{tikz}
\usepackage{mathdots}
\usepackage{yhmath}
\usepackage{cancel}
\usepackage{array}
\usepackage{multirow}
\usepackage{textcomp}
\usepackage{gensymb}
\usepackage{tabularx}
\usepackage{extarrows}
\usetikzlibrary{fadings}
\usetikzlibrary{patterns}
\usetikzlibrary{shadows.blur}
\usetikzlibrary{shapes}

\newcolumntype{?}{!{\vrule width 1pt}}

\newcommand{\SO}{\ensuremath{\mathsf{SO(3)}}}

\renewcommand{\Re}{\ensuremath{\mathbb{R}}}

\title{\LARGE \bf
	Hardware- and Vision-in-the-Loop Validation of Deep Monocular Pose Estimation for Autonomous Maritime UAV Flight
}
\author{Maneesha Wickramasuriya, Beomyeol Yu, Jaden Shin, Mason Huslig, Taeyoung Lee, and Murray Snyder
	\thanks{Maneesha Wickramasuriya, Beomyeol Yu, Mason Huslig, Jaden Shin, Taeyoung Lee, and Murray Snyder, Mechanical and Aerospace Engineering, George Washington University, Washington, DC 20052, {\tt \{maneesh,yubeomyeol,jaden.shin,mason.huslig, \ tylee,snydermr\}@gwu.edu}}%
	\thanks{\textsuperscript{\footnotesize\ensuremath{*}}This research has been supported in part by NAWCAD (N004212610005), USNA/NAVSUP (N0016123RC01EA5), AFOSR MURI (FA9550-23-1-0400), and ONR (N00014-23-1-2850).}
}

\graphicspath{{./figs/}}

\begin{document}
	\allowdisplaybreaks
	\maketitle \thispagestyle{empty} \pagestyle{empty}
	
	\begin{abstract}
		Autonomous UAV operations on ships require reliable vision-based relative pose estimation, yet at-sea validation is costly, weather-dependent, and risky.
		This paper presents a hardware-validated vision-in-the-loop framework that enables fully autonomous indoor flight while emulating photorealistic maritime environments.
		Rendered maritime views are processed onboard by a deep transformer-based monocular pose estimator.
		Delayed vision measurements are fused with high-rate IMU data using a delayed Kalman filter to provide consistent state estimates for geometric control.
		The system captures critical embedded effects, including perception latency, asynchronous updates, and computational constraints, that are absent in pure simulation.
		Autonomous takeoff, trajectory tracking, and landing experiments demonstrate stable closed-loop flight.
		The results establish a safe and hardware-realistic intermediate stage for developing maritime UAV autonomy prior to shipboard deployment.
	\end{abstract}

	\section{Introduction}
	
	Accurate estimation of a UAV’s pose relative to a moving ship is essential for safe takeoff, trajectory tracking, and landing in maritime environments~\cite{JGCD_Ship_Airwakes,Bostock}.
	Vision-based approaches provide a promising solution for ship-relative 6D pose estimation without relying on external infrastructure such as GPS.
	
	Our prior work introduced a deep monocular pose estimation method based on a Transformer Neural Network Multi-Object (TNN-MO) architecture~\cite{wick2025JGCD}.
	The method estimates ship-relative 6D pose from a single RGB image by leveraging geometric structure and multi-object keypoint reasoning.
	
	Despite advances in deep vision-based pose estimation, real-ocean validation remains costly, weather-dependent, and risky~\cite{Bostock2025_JIRS}.
	To address this, we previously proposed a vision-in-the-loop framework that renders photorealistic maritime imagery during indoor flight using 3D Gaussian Splatting (3DGS)~\cite{wick2025icuas,kerbl3Dgaussians}.
	Extending it to closed-loop autonomous flight introduces practical challenges absent in simulation, including latency, irregular measurements, and estimator--controller coupling.
	These effects motivate hardware-validated experiments under realistic embedded conditions.
	
	This paper presents a hardware-validated closed-loop autonomy framework that bridges perception development and maritime deployment.
	It integrates photorealistic maritime rendering, onboard deep transformer inference, delay-aware state estimation, and geometric control on a flying UAV.
	During flight, maritime views rendered from a 3DGS model using motion-capture pose are streamed to onboard edge computing, where TNN-MO estimates ship-relative 6D pose.
	To maintain real-time control under latency and asynchronous updates, delayed TNN-MO measurements are fused with high-rate IMU data using a delayed Kalman filter (DKF), providing consistent current-time state estimates for geometric control~\cite{LeeLeoPICDC10}.
	Executing the full perception--estimation--control stack on embedded hardware exposes practical deployment effects, including end-to-end delay, timing variability, and estimator--controller interaction.
	
	The primary contributions of this paper are as follows.
	First, we integrate photorealistic maritime rendering, onboard deep monocular pose estimation, delay-aware state estimation, and geometric control on a single embedded UAV, enabling fully autonomous closed-loop indoor flight.
	Second, we characterize and compensate for end-to-end perception latency, showing that a delayed Kalman filter maintains estimator consistency under asynchronous and out-of-sequence measurements~\cite{Tasoulis2010}.
	Third, we verify that delayed monocular vision can support stable real-time autonomous flight, providing a practical step between simulation and shipboard deployment.
	
	The proposed framework is expected to accelerate and de-risk maritime UAV autonomy development by reducing reliance on costly, weather-constrained sea trials while enabling systematic, hardware-realistic validation of perception, estimation, and control under embedded operating conditions.
	
	\section{Hardware-Validated Vision-in-the-Loop Autonomy Framework}
	
	This section presents the complete hardware-validated autonomy architecture that integrates maritime scene emulation, deep transformer-based perception, delay-aware state estimation, and geometric control within a physically flying UAV operating under realistic embedded and communication constraints.
	The framework is designed to bridge simulation-based perception development and real maritime deployment by validating the full perception–estimation–control loop in hardware.
	
	This section outlines the overall system architecture and its core components: the photorealistic virtual environment, deep monocular pose estimator, and delayed Kalman filter.
	
	\subsection{System Architecture and Data Flow}
	
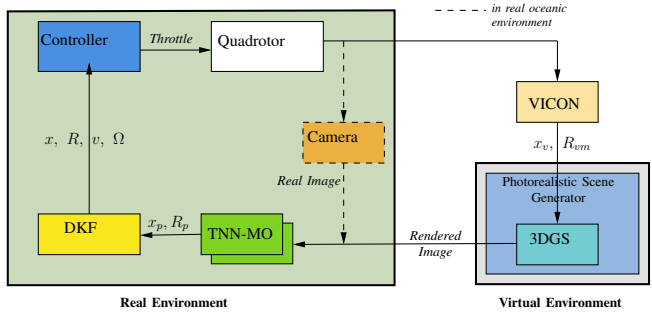
\begin{figure}[t]
		\centering
		\resizebox{\linewidth}{!}{%
			\let\origfontfamily\fontfamily
			\def\fontfamily#1{\origfontfamily{cmr}}
			
			\tikzset{every picture/.style={line width=0.75pt}} 
			
			\begin{tikzpicture}[x=0.75pt,y=0.75pt,yscale=-1,xscale=1]
				
				\draw  [fill={rgb, 255:red, 65; green, 117; blue, 5 }  ,fill opacity=0.27 ][line width=1.5]  (20,61) -- (202.91,60.52) -- (400,60) -- (400,140) -- (400,330) -- (170,330) -- (20,330) -- (20,225.67) -- cycle ;
				\draw  [fill={rgb, 255:red, 155; green, 155; blue, 155 }  ,fill opacity=0.3 ][line width=1.5]  (480,210) -- (650,210) -- (650,329) -- (480,329) -- cycle ;
				\draw  [fill={rgb, 255:red, 74; green, 144; blue, 226 }  ,fill opacity=0.51 ] (490,220) -- (640,220) -- (640,319) -- (490,319) -- cycle ;
				\draw  [fill={rgb, 255:red, 255; green, 255; blue, 255 }  ,fill opacity=1 ] (220,71) -- (330,71) -- (330,120) -- (220,120) -- cycle ;
				\draw  [fill={rgb, 255:red, 74; green, 144; blue, 226 }  ,fill opacity=1 ] (50,71) -- (150,71) -- (150,120) -- (50,120) -- cycle ;
				\draw  [fill={rgb, 255:red, 248; green, 231; blue, 28 }  ,fill opacity=1 ] (50,260) -- (150,260) -- (150,300) -- (50,300) -- cycle ;
				\draw  [fill={rgb, 255:red, 248; green, 189; blue, 28 }  ,fill opacity=0.42 ] (520,130) -- (600,130) -- (600,170) -- (520,170) -- cycle ;
				\draw    (560,90) -- (330,90) ;
				\draw    (150,99) -- (218,99) ;
				\draw [shift={(220,99)}, rotate = 180] [fill={rgb, 255:red, 0; green, 0; blue, 0 }  ][line width=0.08]  [draw opacity=0] (12,-3) -- (0,0) -- (12,3) -- cycle    ;
				\draw    (520,289) -- (336,289.77) -- (302,289.99) ;
				\draw [shift={(300,290)}, rotate = 359.63] [fill={rgb, 255:red, 0; green, 0; blue, 0 }  ][line width=0.08]  [draw opacity=0] (12,-3) -- (0,0) -- (12,3) -- cycle    ;
				\draw    (100,260) -- (100,113) ;
				\draw [shift={(100,111)}, rotate = 90] [fill={rgb, 255:red, 0; green, 0; blue, 0 }  ][line width=0.08]  [draw opacity=0] (12,-3) -- (0,0) -- (12,3) -- cycle    ;
				\draw  [fill={rgb, 255:red, 126; green, 211; blue, 33 }  ,fill opacity=1 ][line width=0.75]  (220,269) -- (300,269) -- (300,309) -- (220,309) -- cycle ;
				\draw  [fill={rgb, 255:red, 245; green, 166; blue, 35 }  ,fill opacity=0.8 ][dash pattern={on 4.5pt off 4.5pt}] (310,170) -- (390,170) -- (390,210) -- (310,210) -- cycle ;
				\draw  [fill={rgb, 255:red, 80; green, 227; blue, 194 }  ,fill opacity=0.46 ] (520,270) -- (600,270) -- (600,310) -- (520,310) -- cycle ;
				\draw  [dash pattern={on 4.5pt off 4.5pt}]  (440,60) -- (471,60) -- (478.23,60) -- (490,60) ;
				\draw    (210,280) -- (152,280.97) ;
				\draw [shift={(150,281)}, rotate = 359.05] [fill={rgb, 255:red, 0; green, 0; blue, 0 }  ][line width=0.08]  [draw opacity=0] (12,-3) -- (0,0) -- (12,3) -- cycle    ;
				\draw    (560,90) -- (560,128) ;
				\draw [shift={(560,130)}, rotate = 270] [fill={rgb, 255:red, 0; green, 0; blue, 0 }  ][line width=0.08]  [draw opacity=0] (12,-3) -- (0,0) -- (12,3) -- cycle    ;
				\draw  [dash pattern={on 4.5pt off 4.5pt}]  (350,90) -- (350,168) ;
				\draw [shift={(350,170)}, rotate = 270] [fill={rgb, 255:red, 0; green, 0; blue, 0 }  ][line width=0.08]  [draw opacity=0] (12,-3) -- (0,0) -- (12,3) -- cycle    ;
				\draw  [dash pattern={on 4.5pt off 4.5pt}]  (350,210) -- (350,288) ;
				\draw [shift={(350,290)}, rotate = 270] [fill={rgb, 255:red, 0; green, 0; blue, 0 }  ][line width=0.08]  [draw opacity=0] (12,-3) -- (0,0) -- (12,3) -- cycle    ;
				\draw    (560,170) -- (560,268) ;
				\draw [shift={(560,270)}, rotate = 270] [fill={rgb, 255:red, 0; green, 0; blue, 0 }  ][line width=0.08]  [draw opacity=0] (12,-3) -- (0,0) -- (12,3) -- cycle    ;
				\draw  [fill={rgb, 255:red, 126; green, 211; blue, 33 }  ,fill opacity=1 ][line width=0.75]  (210,260) -- (290,260) -- (290,300) -- (210,300) -- cycle ;
				
				\draw (225,83) node [anchor=north west][inner sep=0.75pt]  [font=\large] [align=left] {{\large Quadrotor}};
				\draw (51,82) node [anchor=north west][inner sep=0.75pt]  [font=\large] [align=left] {{\large Controller}};
				\draw (74,267) node [anchor=north west][inner sep=0.75pt]  [font=\large] [align=left] {{\large DKF}};
				\draw (530,144) node [anchor=north west][inner sep=0.75pt]  [font=\large] [align=left] {VICON};
				\draw (55,178.4) node [anchor=north west][inner sep=0.75pt]  [font=\large]  {$x,\ R,\ v,\ \Omega $};
				\draw (158,81) node [anchor=north west][inner sep=0.75pt]  [font=\normalsize] [align=left] {\textit{Throttle }};
				\draw (535,182.4) node [anchor=north west][inner sep=0.75pt]  [font=\large]  {$x_v,\ R_{vm}$};
				\draw (215,272) node [anchor=north west][inner sep=0.75pt]  [font=\large] [align=left] {TNN-MO};
				\draw (156,260) node [anchor=north west][inner sep=0.75pt]  [font=\large]  {$x_p ,R_p$};
				\draw (531,277) node [anchor=north west][inner sep=0.75pt]  [font=\large] [align=left] {3DGS};
				\draw (411,275) node [anchor=north west][inner sep=0.75pt]  [font=\normalsize] [align=left] {\begin{minipage}[lt]{41.85pt}\setlength\topsep{0pt}
						\begin{center}
							\textit{Rendered}\\\textit{Image}
						\end{center}
						
				\end{minipage}};
				\draw (492,222) node [anchor=north west][inner sep=0.75pt]  [font=\normalsize] [align=left] {\begin{minipage}[lt]{101.37pt}\setlength\topsep{0pt}
						\begin{center}
							Photorealistic Scene \\Generator
						\end{center}
						
				\end{minipage}};
				\draw (501,341) node [anchor=north west][inner sep=0.75pt]   [align=left] {\textbf{Virtual Environment}};
				\draw (129,341) node [anchor=north west][inner sep=0.75pt]   [align=left] {\textbf{Real Environment}};
				\draw (278,221) node [anchor=north west][inner sep=0.75pt]  [font=\normalsize] [align=left] {\begin{minipage}[lt]{54.88pt}\setlength\topsep{0pt}
						\begin{center}
							\textit{Real Image}
						\end{center}
						
				\end{minipage}};
				\draw (313,177) node [anchor=north west][inner sep=0.75pt]  [font=\large] [align=left] {{\large Camera}};
				\draw (494,52) node [anchor=north west][inner sep=0.75pt]  [font=\small] [align=left] {\textit{in real oceanic  }\\\textit{environment }};
				\draw (40,12) node [anchor=north west][inner sep=0.75pt]  [font=\LARGE,color={rgb, 255:red, 0; green, 0; blue, 0 }  ,opacity=1 ] [align=left] {};

			\end{tikzpicture}

		}%
		
		\caption{Hardware-validated vision-in-the-loop simulation framework for autonomous indoor UAV flight, illustrating parallel TNN-MO inference and DKF-based state estimation for closed-loop control on an NVIDIA Jetson Orin NX.
			During indoor experiments, the onboard camera is not used; instead, the perception stack consumes rendered ``at-sea'' frames, while the physical camera is reserved for future real-ocean deployments.}
		\label{fig:VILAF}
	\end{figure}

	The overall system architecture is illustrated in \Cref{fig:VILAF}.
	The framework enables fully autonomous indoor flight while emulating maritime shipboard operations through real-time photorealistic scene rendering.
	
	During flight, the UAV pose is measured by a Vicon motion-capture system at 200\,Hz and transmitted to an external rendering workstation.
	Using the measured pose, the renderer synthesizes photorealistic maritime views from a 3D Gaussian Splatting (3DGS) scene model in real time.
	The rendered images are streamed over WiFi to an onboard NVIDIA Jetson Orin NX for perception processing.
	
	Onboard, the Transformer Neural Network Multi-Object (TNN-MO) model estimates the ship-relative 6D pose from the rendered imagery.
	Because rendering, communication, and embedded inference introduce non-negligible latency of 0.3 seconds, the resulting vision measurements are time-delayed and asynchronous.
	These delayed pose estimates are fused with high-rate IMU data using a delayed Kalman filter (DKF), which applies measurement correction at the appropriate historical timestamp and repropagates the state forward to recover a consistent current-time estimate.
	The estimated state is then provided to a geometric controller that computes rotor thrust commands for closed-loop autonomous flight.
	
	Unlike pure simulation, this architecture exposes critical system-level effects, including end-to-end perception latency, irregular update timing, GPU/CPU resource contention, communication variability, and estimator–controller coupling.
	By operating the full stack on embedded hardware during physical flight, the framework enables realistic validation of maritime UAV autonomy under practical deployment constraints.
	
	\subsection{Photorealistic Virtual Environments}\label{sec:PRVE}
	
	Realistic evaluation of vision-based autonomy for shipboard UAV operations is costly and risky to conduct at sea due to weather dependence and the potential for hardware loss.
	Building on our prior vision-in-the-loop framework~\cite{wick2025icuas}, we emulate maritime operations during indoor flight by rendering photorealistic ``at-sea'' camera views from the UAV's tracked pose and streaming them to the onboard perception stack.
	
	The virtual environment is constructed using 3D Gaussian Splatting (3DGS)~\cite{kerbl3Dgaussians}, which represents a scene as optimized anisotropic 3D Gaussians and enables efficient, high-quality novel-view synthesis.
	The 3DGS model is generated from multi-view imagery collected around a research vessel at sea, capturing the ship, deck, and surrounding ocean and sky.
	The resulting compact scene representation supports real-time rendering at $640 \times 480$ resolution and 60--110~FPS on an RTX 3060 laptop.
	Although dynamic water regions may appear slightly blurred, structural fidelity of the ship and landing zone is prioritized.
	Domain randomization is applied to reduce sensitivity to variations in ocean and sky appearance~\cite{wick2024conf}, yielding imagery suitable for robust monocular pose estimation.
	
	\subsection{Deep Monocular Pose Estimation}\label{sec:TNN}
	
	Given a single RGB frame generated by the virtual environment, the TNN-MO (Transformer Neural Network Multi-Object) model estimates the ship-relative 6D pose by detecting multiple ship components and fusing their individual pose estimates~\cite{wick2025JGCD}.
	
	\paragraph{Network Architecture}
	
	TNN-MO combines (i) a ResNet50 CNN backbone, (ii) a transformer encoder--decoder, and (iii) probabilistic fusion across detected objects~\cite{wick2025JGCD}.
	An input image ($480\times640$) is encoded into a compact feature map and processed by a 6-layer transformer encoder.
	A 6-layer decoder employs seven learned queries (six ship components plus a ``no-object'' token) to predict object presence and 2D keypoints.
	Object-wise 6D pose estimates are recovered via EPnP~\cite{EPnP}, and Bayesian fusion produces the final pose estimate with confidence-based weighting~\cite{wick2025JGCD}.
	
	\paragraph{Training and Validation}
	
	TNN-MO is trained primarily in simulation using large-scale synthetic ship-based data with domain randomization over textures, lighting, weather, and camera viewpoints to improve generalization~\cite{wick2024conf,wick2025JGCD}.
	It is evaluated on synthetic, real-world shipboard, and photorealistic 3DGS-rendered data, where it consistently provides accurate and robust pose estimates under varying conditions~\cite{wick2025JGCD,wick2025icuas}.
	
	\subsection{Sensor Fusion with Delayed Kalman Filter}\label{sec:DKF}
	
	TNN-MO pose inference introduces non-negligible latency on embedded hardware. To provide high-rate current-time state estimates for control, we use a delayed Kalman filter (DKF)~\cite{Gamagedara2019} that fuses delayed TNN-MO pose measurements with high-rate IMU data.
	
	The estimated state consists of attitude $R\in\SO$, position and velocity $x,v\in\Re^3$, and accelerometer bias $b_a\in\Re^3$. The DKF first propagates this state and its covariance using IMU measurements. When a delayed TNN-MO pose measurement arrives, the filter applies the update at the corresponding past measurement time using a history buffer of stored states, covariances, and IMU data.
	
	The corrected state is then re-propagated to the current time, yielding delay-compensated estimates suitable for feedback control. This enables consistent state estimation despite latency, asynchronous updates, and out-of-sequence measurements.
	
	
	\begin{table*}[htbp]
		\centering
		\caption{TNN-MO inference performance on Jetson Orin NX.}
		\label{tab:tnn-performance}
		\begin{tabular}{l *{8}{c}}
			\toprule
			& \multicolumn{2}{c}{\textbf{Single}} & \multicolumn{2}{c}{\textbf{Two}} & \multicolumn{2}{c}{\textbf{Three}} & \multicolumn{2}{c}{\textbf{Four}} \\
			\cmidrule(lr){2-3} \cmidrule(lr){4-5} \cmidrule(lr){6-7} \cmidrule(lr){8-9}
			& w/o FC & w/ FC & w/o FC & w/ FC & w/o FC & w/ FC & w/o FC & w/ FC  \\
			\midrule
			Avg. Frequency (Hz)   & 5.5  & 4.7  & 9.0  & 8.7  & 9.8  & 9.0  & 11.1 & 10.0 \\
			Avg. delay (s)   & 0.18 & 0.21 & 0.22 & 0.23 & 0.30 & 0.33 & 0.36 & 0.40 \\
			\bottomrule
		\end{tabular}
	\end{table*}
	
	\subsection{Geometric Control}
	
	Finally, the quadrotor is controlled using a geometric tracking controller~\cite{LeeLeoPICDC10}.
	The controller computes total thrust and body moments to track desired position and attitude trajectories without relying on local attitude parameterizations.
	Using the current-time state estimate provided by the DKF, the controller maintains stable closed-loop flight despite delayed and asynchronous vision measurements.
	This formulation ensures consistent tracking performance under realistic embedded operating conditions.

	\section{Embedded Implementation and Experimental Realization}\label{sec:implementation}
	
	This section describes the real-time embedded realization of the proposed autonomy framework and characterizes system-level effects that influence closed-loop performance, including resource constraints, latency variability, and scheduling trade-offs.
	
	\subsection{Experimental Architecture in Indoor Maritime Emulation}
	
	In a real shipboard mission, the onboard camera directly observes the maritime scene and TNN-MO estimates the ship-relative pose.
	In the proposed vision-in-the-loop implementation (\Cref{fig:VILAF}), the UAV instead flies indoors while maritime imagery is emulated through real-time rendering.
	
	For indoor experiments, the UAV pose is measured at 200~Hz using a Vicon motion capture system with twelve Valkyrie VK-8 cameras, achieving sub-millimeter accuracy under optimal conditions.
	Retro-reflective markers mounted on the UAV (see \Cref{fig:UAV}) enable estimation of the marker frame pose.
	This pose is transmitted via Ethernet to an external Linux workstation equipped with an RTX 2080 GPU, where photorealistic maritime views are rendered from the 3DGS scene model.
	The rendered images are streamed over Wi-Fi to the onboard Jetson Orin NX for perception processing.
	
	Onboard, multiple TNN-MO instances process the rendered images to produce delayed pose measurements.
	These measurements are fused with 200~Hz IMU data from a VectorNav VN-100 to recover the current-time state estimate.
	The estimated state is then supplied to the geometric controller, which computes rotor thrust commands for closed-loop autonomous flight while the UAV ``perceives'' the virtual maritime environment.
	
	\subsection{Parallel Onboard Inference and Latency Characterization}\label{sec:Onboard}

	\begin{figure}[b]
		\centering
		\resizebox{\linewidth}{!}{%
			\let\origfontfamily\fontfamily
			\def\fontfamily#1{\origfontfamily{cmr}}
			\tikzset{every picture/.style={line width=0.75pt}} 
			
			\begin{tikzpicture}[x=0.75pt,y=0.75pt,yscale=-1,xscale=1]
				
				\draw  [draw opacity=0][fill={rgb, 255:red, 126; green, 211; blue, 33 }  ,fill opacity=1 ] (250,100) -- (380,100) -- (380,120) -- (250,120) -- cycle ;
				\draw  [draw opacity=0][fill={rgb, 255:red, 184; green, 233; blue, 134 }  ,fill opacity=1 ] (380,100) -- (400,100) -- (400,120) -- (380,120) -- cycle ;
				\draw    (400,110) -- (400,82) ;
				\draw [shift={(400,80)}, rotate = 90] [color={rgb, 255:red, 0; green, 0; blue, 0 }  ][line width=0.75]    (10.93,-3.29) .. controls (6.95,-1.4) and (3.31,-0.3) .. (0,0) .. controls (3.31,0.3) and (6.95,1.4) .. (10.93,3.29)   ;
				\draw  [draw opacity=0][fill={rgb, 255:red, 126; green, 211; blue, 33 }  ,fill opacity=1 ] (300,130) -- (430,130) -- (430,150) -- (300,150) -- cycle ;
				\draw  [draw opacity=0][fill={rgb, 255:red, 184; green, 233; blue, 134 }  ,fill opacity=1 ] (430,130) -- (450,130) -- (450,150) -- (430,150) -- cycle ;
				\draw    (450,140) -- (450,112) ;
				\draw [shift={(450,110)}, rotate = 90] [color={rgb, 255:red, 0; green, 0; blue, 0 }  ][line width=0.75]    (10.93,-3.29) .. controls (6.95,-1.4) and (3.31,-0.3) .. (0,0) .. controls (3.31,0.3) and (6.95,1.4) .. (10.93,3.29)   ;
				\draw  [draw opacity=0][fill={rgb, 255:red, 126; green, 211; blue, 33 }  ,fill opacity=1 ] (350,160) -- (480,160) -- (480,180) -- (350,180) -- cycle ;
				\draw  [draw opacity=0][fill={rgb, 255:red, 184; green, 233; blue, 134 }  ,fill opacity=1 ] (480,160) -- (500,160) -- (500,180) -- (480,180) -- cycle ;
				\draw    (500,170) -- (500,142) ;
				\draw [shift={(500,140)}, rotate = 90] [color={rgb, 255:red, 0; green, 0; blue, 0 }  ][line width=0.75]    (10.93,-3.29) .. controls (6.95,-1.4) and (3.31,-0.3) .. (0,0) .. controls (3.31,0.3) and (6.95,1.4) .. (10.93,3.29)   ;
				\draw  [draw opacity=0][fill={rgb, 255:red, 126; green, 211; blue, 33 }  ,fill opacity=1 ] (200,70) -- (330,70) -- (330,90) -- (200,90) -- cycle ;
				\draw  [draw opacity=0][fill={rgb, 255:red, 80; green, 227; blue, 194 }  ,fill opacity=1 ] (180,70) -- (200,70) -- (200,90) -- (180,90) -- cycle ;
				\draw  [draw opacity=0][fill={rgb, 255:red, 184; green, 233; blue, 134 }  ,fill opacity=1 ] (330,70) -- (350,70) -- (350,90) -- (330,90) -- cycle ;
				\draw    (350,80) -- (350,52) ;
				\draw [shift={(350,50)}, rotate = 90] [color={rgb, 255:red, 0; green, 0; blue, 0 }  ][line width=0.75]    (10.93,-3.29) .. controls (6.95,-1.4) and (3.31,-0.3) .. (0,0) .. controls (3.31,0.3) and (6.95,1.4) .. (10.93,3.29)   ;
				\draw  [draw opacity=0][fill={rgb, 255:red, 74; green, 144; blue, 226 }  ,fill opacity=1 ] (160,70) -- (180,70) -- (180,90) -- (160,90) -- cycle ;
				\draw  [draw opacity=0][fill={rgb, 255:red, 245; green, 166; blue, 35 }  ,fill opacity=1 ] (150,70) -- (160,70) -- (160,90) -- (150,90) -- cycle ;
				\draw  [draw opacity=0][fill={rgb, 255:red, 80; green, 227; blue, 194 }  ,fill opacity=1 ] (230,100) -- (250,100) -- (250,120) -- (230,120) -- cycle ;
				\draw  [draw opacity=0][fill={rgb, 255:red, 74; green, 144; blue, 226 }  ,fill opacity=1 ] (210,100) -- (230,100) -- (230,120) -- (210,120) -- cycle ;
				\draw  [draw opacity=0][fill={rgb, 255:red, 245; green, 166; blue, 35 }  ,fill opacity=1 ] (200,100) -- (210,100) -- (210,120) -- (200,120) -- cycle ;
				\draw  [draw opacity=0][fill={rgb, 255:red, 80; green, 227; blue, 194 }  ,fill opacity=1 ] (280,130) -- (300,130) -- (300,150) -- (280,150) -- cycle ;
				\draw  [draw opacity=0][fill={rgb, 255:red, 74; green, 144; blue, 226 }  ,fill opacity=1 ] (260,130) -- (280,130) -- (280,150) -- (260,150) -- cycle ;
				\draw  [draw opacity=0][fill={rgb, 255:red, 245; green, 166; blue, 35 }  ,fill opacity=1 ] (250,130) -- (260,130) -- (260,150) -- (250,150) -- cycle ;
				\draw  [draw opacity=0][fill={rgb, 255:red, 80; green, 227; blue, 194 }  ,fill opacity=1 ] (330,160) -- (350,160) -- (350,180) -- (330,180) -- cycle ;
				\draw  [draw opacity=0][fill={rgb, 255:red, 74; green, 144; blue, 226 }  ,fill opacity=1 ] (310,160) -- (330,160) -- (330,180) -- (310,180) -- cycle ;
				\draw  [draw opacity=0][fill={rgb, 255:red, 245; green, 166; blue, 35 }  ,fill opacity=1 ] (300,160) -- (310,160) -- (310,180) -- (300,180) -- cycle ;
				\draw  [dash pattern={on 4.5pt off 4.5pt}]  (140,100) -- (400,100) ;
				\draw  [dash pattern={on 4.5pt off 4.5pt}]  (140,150) -- (500,150) ;
				\draw    (402,50) -- (448,50) ;
				\draw [shift={(450,50)}, rotate = 180] [color={rgb, 255:red, 0; green, 0; blue, 0 }  ][line width=0.75]    (10.93,-3.29) .. controls (6.95,-1.4) and (3.31,-0.3) .. (0,0) .. controls (3.31,0.3) and (6.95,1.4) .. (10.93,3.29)   ;
				\draw [shift={(400,50)}, rotate = 0] [color={rgb, 255:red, 0; green, 0; blue, 0 }  ][line width=0.75]    (10.93,-3.29) .. controls (6.95,-1.4) and (3.31,-0.3) .. (0,0) .. controls (3.31,0.3) and (6.95,1.4) .. (10.93,3.29)   ;
				\draw  [dash pattern={on 0.84pt off 2.51pt}]  (400,50) -- (400,80) ;
				\draw  [dash pattern={on 0.84pt off 2.51pt}]  (450,50) -- (450,110) ;
				\draw    (302,215) -- (498,215) ;
				\draw [shift={(500,215)}, rotate = 180] [color={rgb, 255:red, 0; green, 0; blue, 0 }  ][line width=0.75]    (10.93,-3.29) .. controls (6.95,-1.4) and (3.31,-0.3) .. (0,0) .. controls (3.31,0.3) and (6.95,1.4) .. (10.93,3.29)   ;
				\draw [shift={(300,215)}, rotate = 0] [color={rgb, 255:red, 0; green, 0; blue, 0 }  ][line width=0.75]    (10.93,-3.29) .. controls (6.95,-1.4) and (3.31,-0.3) .. (0,0) .. controls (3.31,0.3) and (6.95,1.4) .. (10.93,3.29)   ;
				\draw    (352,195) -- (498,195) ;
				\draw [shift={(500,195)}, rotate = 180] [color={rgb, 255:red, 0; green, 0; blue, 0 }  ][line width=0.75]    (10.93,-3.29) .. controls (6.95,-1.4) and (3.31,-0.3) .. (0,0) .. controls (3.31,0.3) and (6.95,1.4) .. (10.93,3.29)   ;
				\draw [shift={(350,195)}, rotate = 0] [color={rgb, 255:red, 0; green, 0; blue, 0 }  ][line width=0.75]    (10.93,-3.29) .. controls (6.95,-1.4) and (3.31,-0.3) .. (0,0) .. controls (3.31,0.3) and (6.95,1.4) .. (10.93,3.29)   ;
				\draw    (140,102) -- (140,148) ;
				\draw [shift={(140,150)}, rotate = 270] [color={rgb, 255:red, 0; green, 0; blue, 0 }  ][line width=0.75]    (10.93,-3.29) .. controls (6.95,-1.4) and (3.31,-0.3) .. (0,0) .. controls (3.31,0.3) and (6.95,1.4) .. (10.93,3.29)   ;
				\draw [shift={(140,100)}, rotate = 90] [color={rgb, 255:red, 0; green, 0; blue, 0 }  ][line width=0.75]    (10.93,-3.29) .. controls (6.95,-1.4) and (3.31,-0.3) .. (0,0) .. controls (3.31,0.3) and (6.95,1.4) .. (10.93,3.29)   ;
				\draw  [draw opacity=0][fill={rgb, 255:red, 245; green, 166; blue, 35 }  ,fill opacity=1 ] (50,275) -- (60,275) -- (60,285) -- (50,285) -- cycle ;
				\draw  [draw opacity=0][fill={rgb, 255:red, 74; green, 144; blue, 226 }  ,fill opacity=1 ] (50,295) -- (60,295) -- (60,305) -- (50,305) -- cycle ;
				\draw  [draw opacity=0][fill={rgb, 255:red, 80; green, 227; blue, 194 }  ,fill opacity=1 ] (50,315) -- (60,315) -- (60,325) -- (50,325) -- cycle ;
				\draw  [draw opacity=0][fill={rgb, 255:red, 126; green, 211; blue, 33 }  ,fill opacity=1 ] (290,275) -- (300,275) -- (300,285) -- (290,285) -- cycle ;
				\draw  [draw opacity=0][fill={rgb, 255:red, 184; green, 233; blue, 134 }  ,fill opacity=1 ] (290,295) -- (300,295) -- (300,305) -- (290,305) -- cycle ;
				\draw    (20,240) -- (548,240) ;
				\draw [shift={(550,240)}, rotate = 180] [color={rgb, 255:red, 0; green, 0; blue, 0 }  ][line width=0.75]    (10.93,-3.29) .. controls (6.95,-1.4) and (3.31,-0.3) .. (0,0) .. controls (3.31,0.3) and (6.95,1.4) .. (10.93,3.29)   ;
				\draw  [draw opacity=0][fill={rgb, 255:red, 255; green, 255; blue, 255 }  ,fill opacity=0.48 ] (120,70) -- (350,70) -- (350,90) -- (120,90) -- cycle ;
				\draw  [draw opacity=0][fill={rgb, 255:red, 255; green, 255; blue, 255 }  ,fill opacity=0.48 ] (270,160) -- (500,160) -- (500,180) -- (270,180) -- cycle ;
				\draw  [draw opacity=0][fill={rgb, 255:red, 126; green, 211; blue, 33 }  ,fill opacity=1 ] (150,40) -- (280,40) -- (280,60) -- (150,60) -- cycle ;
				\draw  [draw opacity=0][fill={rgb, 255:red, 184; green, 233; blue, 134 }  ,fill opacity=1 ] (280,40) -- (300,40) -- (300,60) -- (280,60) -- cycle ;
				\draw    (300,50) -- (300,22) ;
				\draw [shift={(300,20)}, rotate = 90] [color={rgb, 255:red, 0; green, 0; blue, 0 }  ][line width=0.75]    (10.93,-3.29) .. controls (6.95,-1.4) and (3.31,-0.3) .. (0,0) .. controls (3.31,0.3) and (6.95,1.4) .. (10.93,3.29)   ;
				\draw  [draw opacity=0][fill={rgb, 255:red, 80; green, 227; blue, 194 }  ,fill opacity=1 ] (130,40) -- (150,40) -- (150,60) -- (130,60) -- cycle ;
				\draw  [draw opacity=0][fill={rgb, 255:red, 74; green, 144; blue, 226 }  ,fill opacity=1 ] (110,40) -- (130,40) -- (130,60) -- (110,60) -- cycle ;
				\draw  [draw opacity=0][fill={rgb, 255:red, 245; green, 166; blue, 35 }  ,fill opacity=1 ] (100,40) -- (110,40) -- (110,60) -- (100,60) -- cycle ;
				\draw  [draw opacity=0][fill={rgb, 255:red, 255; green, 255; blue, 255 }  ,fill opacity=0.48 ] (70,40) -- (300,40) -- (300,60) -- (70,60) -- cycle ;
				
				\draw (310,107.5) node   [align=left] {\begin{minipage}[lt]{68pt}\setlength\topsep{0pt}
						{\fontfamily{pcr}\selectfont {\small TNN-MO - 1}}
				\end{minipage}};
				\draw (450,67.5) node   [align=left] {\begin{minipage}[lt]{68pt}\setlength\topsep{0pt}
						{\fontfamily{pcr}\selectfont {\small pose}}
				\end{minipage}};
				\draw (500,97.5) node   [align=left] {\begin{minipage}[lt]{68pt}\setlength\topsep{0pt}
						{\fontfamily{pcr}\selectfont {\small pose}}
				\end{minipage}};
				\draw (550,127.5) node   [align=left] {\begin{minipage}[lt]{68pt}\setlength\topsep{0pt}
						{\fontfamily{pcr}\selectfont {\small pose}}
				\end{minipage}};
				\draw (370,37.5) node  [color={rgb, 255:red, 0; green, 0; blue, 0 }  ,opacity=1 ] [align=left] {\begin{minipage}[lt]{27.2pt}\setlength\topsep{0pt}
						{\fontfamily{pcr}\selectfont {\small pose}}
				\end{minipage}};
				\draw (365,140) node   [align=left] {\begin{minipage}[lt]{68pt}\setlength\topsep{0pt}
						{\fontfamily{pcr}\selectfont {\small TNN-MO - 2}}
				\end{minipage}};
				\draw (415,170) node   [align=left] {\begin{minipage}[lt]{68pt}\setlength\topsep{0pt}
						{\fontfamily{pcr}\selectfont {\small TNN-MO - 1}}
				\end{minipage}};
				\draw (260,77.5) node   [align=left] {\begin{minipage}[lt]{68pt}\setlength\topsep{0pt}
						{\fontfamily{pcr}\selectfont {\small TNN-MO - 2}}
				\end{minipage}};
				\draw (445,37.5) node   [align=left] {\begin{minipage}[lt]{47.6pt}\setlength\topsep{0pt}
						{\fontfamily{pcr}\selectfont {\small 0.117s}}
				\end{minipage}};
				\draw (415,227.5) node   [align=left] {\begin{minipage}[lt]{102pt}\setlength\topsep{0pt}
						{\fontfamily{pcr}\selectfont {\small Total delay 0.3s}}
				\end{minipage}};
				\draw (420,202.5) node   [align=left] {\begin{minipage}[lt]{27.2pt}\setlength\topsep{0pt}
						{\fontfamily{pcr}\selectfont {\small 0.235s}}
				\end{minipage}};
				\draw (75,130) node   [align=left] {\begin{minipage}[lt]{88.4pt}\setlength\topsep{0pt}
						{\fontfamily{pcr}\selectfont {\small No. of TNN-MO models loaded per one loop cycle}}
				\end{minipage}};
				\draw (160,277.5) node  [color={rgb, 255:red, 0; green, 0; blue, 0 }  ,opacity=1 ] [align=left] {\begin{minipage}[lt]{122.4pt}\setlength\topsep{0pt}
						{\fontfamily{pcr}\selectfont {\small motion cap latency 0.005 s}}
				\end{minipage}};
				\draw (170,297.5) node  [color={rgb, 255:red, 0; green, 0; blue, 0 }  ,opacity=1 ] [align=left] {\begin{minipage}[lt]{136pt}\setlength\topsep{0pt}
						{\fontfamily{pcr}\selectfont {\small image rendering latency 0.016 s}}
				\end{minipage}};
				\draw (150,317.5) node  [color={rgb, 255:red, 0; green, 0; blue, 0 }  ,opacity=1 ] [align=left] {\begin{minipage}[lt]{108.8pt}\setlength\topsep{0pt}
						{\fontfamily{pcr}\selectfont {\small WiFi latency 0.075 s}}
				\end{minipage}};
				\draw (425,277.5) node  [color={rgb, 255:red, 0; green, 0; blue, 0 }  ,opacity=1 ] [align=left] {\begin{minipage}[lt]{156.4pt}\setlength\topsep{0pt}
						{\fontfamily{pcr}\selectfont {\small TNN-MO process time \ 0.27 $<$ 0.3s}}
				\end{minipage}};
				\draw (390,297.5) node  [color={rgb, 255:red, 0; green, 0; blue, 0 }  ,opacity=1 ] [align=left] {\begin{minipage}[lt]{108.8pt}\setlength\topsep{0pt}
						{\fontfamily{pcr}\selectfont {\small wait time}}
				\end{minipage}};
				\draw (570,237.5) node   [align=left] {\begin{minipage}[lt]{27.2pt}\setlength\topsep{0pt}
						{\fontfamily{pcr}\selectfont {\small time}}
				\end{minipage}};
				\draw (215,50) node   [align=left] {\begin{minipage}[lt]{68pt}\setlength\topsep{0pt}
						{\fontfamily{pcr}\selectfont {\small TNN-MO - 1}}
				\end{minipage}};
				\draw (320,15) node  [color={rgb, 255:red, 0; green, 0; blue, 0 }  ,opacity=1 ] [align=left] {\begin{minipage}[lt]{27.2pt}\setlength\topsep{0pt}
						{\fontfamily{pcr}\selectfont {\small pose}}
				\end{minipage}};

			\end{tikzpicture}
			
		}%
		
		\caption{Parallel, time-shifted TNN-MO pose predicted instances illustrating overlapping processing. Instances are staggered by 0.117~s, enabling pose outputs every 0.117~s (8.5~Hz), despite each taking 0.345~s (WiFi: 0.075~s, TNN-MO: 0.27~s). The cycle repeats with instance 1 starting again after instance 3.}
		\label{fig:TNNLatency}
	\end{figure}
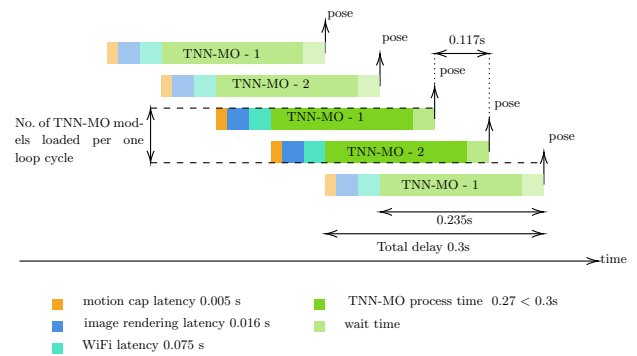

	TNN-MO is deployed on the Jetson Orin NX using a TensorRT-converted FP16 \texttt{.engine} model (instead of the original PyTorch \texttt{.pt} checkpoint) to reduce runtime overhead and enable efficient edge inference.
	As summarized in \Cref{tab:tnn-performance}, a single TNN-MO instance achieves 5.5\,Hz (average inference time 0.18\,s) when the Jetson is dedicated to perception.
	When the flight controller (FC) is also running onboard and competing for CPU resources, the rate decreases to 4.7\,Hz (average inference time 0.21\,s).
	These results demonstrate that achievable update rates are constrained not only by network architecture but also by GPU utilization, CPU scheduling, memory bandwidth, and operating conditions such as temperature and concurrent processes.
	
	To increase effective prediction frequency beyond what a single sequential pipeline allows, multiple TNN-MO instances are executed in parallel, each processing a time-shifted image stream, as illustrated in \Cref{fig:TNNLatency}.
	This pipelined strategy increases throughput because while one instance performs inference, others pre-process or wait for the next frame.
	However, parallelization also increases average measurement delay: each estimate corresponds to an earlier image timestamp due to staggering, and instances contend for shared Jetson resources, which can lengthen inference time.
	Accordingly, \Cref{tab:tnn-performance} shows a clear throughput--latency trade-off, where additional instances increase update frequency but also increase average delay.
	
	In the highest-throughput configuration, four parallel instances with the onboard FC produce approximately 10\,Hz pose updates with an average delay of about 0.4\,s.
	This configuration, however, nearly saturates available GPU memory and compute capacity.
	In addition to inference and scheduling delay, the vision-in-the-loop pipeline includes communication latency of 0.08--0.1\,s, further contributing to total pose-estimation latency observed by the DKF.
	
	To maintain a constant update rate during extended operation, explicit waiting time is inserted (see \Cref{fig:TNNLatency}).
	The commanded rate is set slightly below peak sustained throughput.
	For example, a pipeline initially averaging 8.7\,Hz may gradually decrease to 8.55\,Hz over a two-hour run.
	We therefore target 8.5\,Hz and enforce synchronization at fixed time steps to prevent drift.
	Because the performance of the Jetson is sensitive to ambient temperature, additional tests are conducted under intentionally increased delay conditions.
	Specifically, we evaluate a 0.4\,s inference delay (approximately 0.55\,s total latency including transmission) at a 5\,Hz inference rate to represent a conservative worst-case scenario.

	\subsection{Flight Hardware and Software Integration}
	
	\begin{figure}[b]
		\centering
		\scalebox{0.55}{       
			
			\tikzset{every picture/.style={line width=0.75pt}} 
			
			\begin{tikzpicture}[x=0.75pt,y=0.75pt,yscale=-1,xscale=1]
				
				\draw (378.5,149) node  {\includegraphics[width=303.75pt,height=222pt]{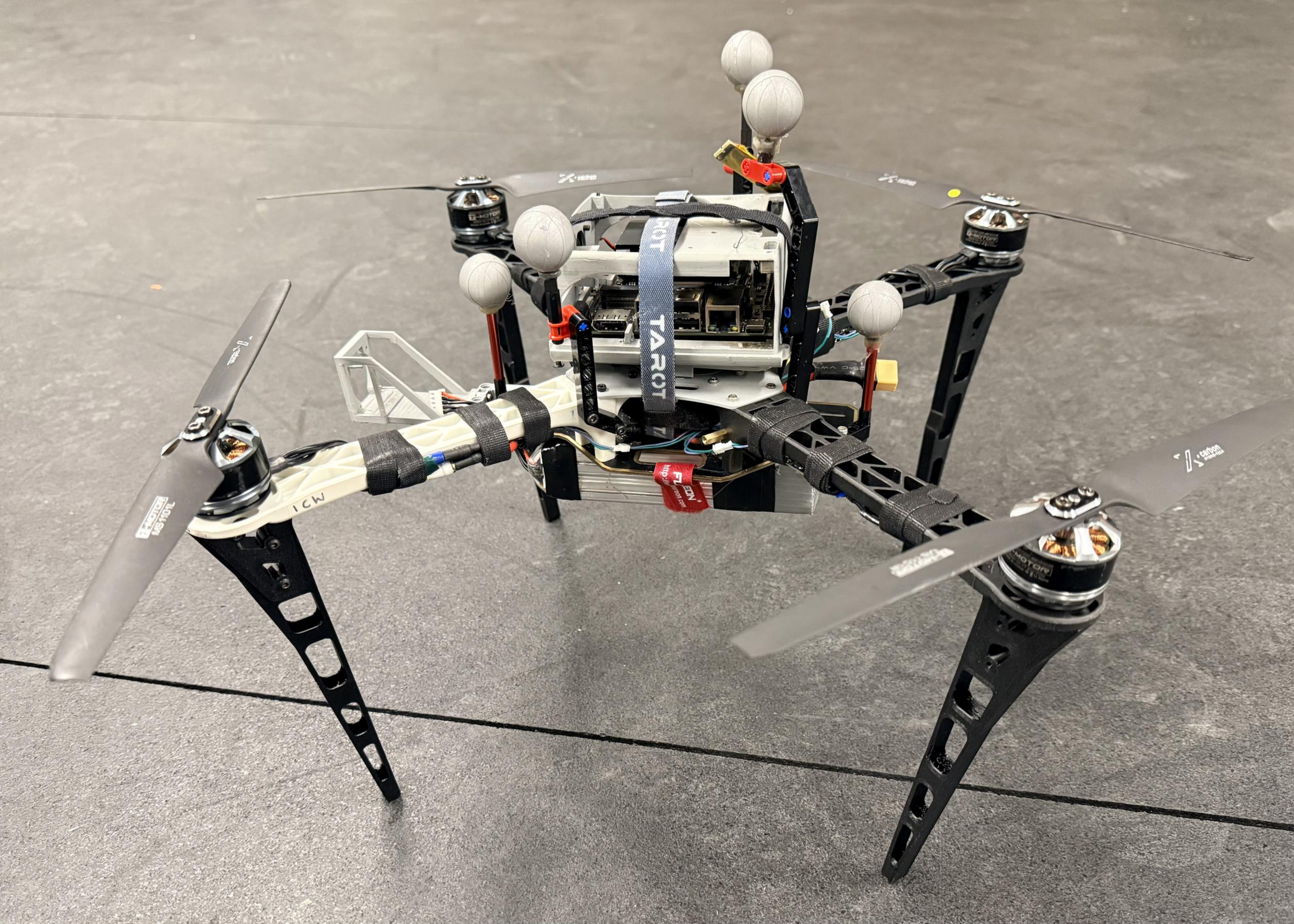}};
				\draw [color={rgb, 255:red, 0; green, 0; blue, 0 }  ,draw opacity=1 ][fill={rgb, 255:red, 255; green, 57; blue, 0 }  ,fill opacity=1 ][line width=3]    (360,29) -- (354.18,58.12) ;
				\draw [shift={(353,64)}, rotate = 281.31] [fill={rgb, 255:red, 0; green, 0; blue, 0 }  ,fill opacity=1 ][line width=0.08]  [draw opacity=0] (18.75,-9.01) -- (0,0) -- (18.75,9.01) -- (12.45,0) -- cycle    ;
				\draw [color={rgb, 255:red, 0; green, 0; blue, 0 }  ,draw opacity=1 ][fill={rgb, 255:red, 255; green, 57; blue, 0 }  ,fill opacity=1 ][line width=3]    (439.92,26.92) -- (448.92,76.1) ;
				\draw [shift={(450,82)}, rotate = 259.63] [fill={rgb, 255:red, 0; green, 0; blue, 0 }  ,fill opacity=1 ][line width=0.08]  [draw opacity=0] (18.75,-9.01) -- (0,0) -- (18.75,9.01) -- (12.45,0) -- cycle    ;
				
				\draw (190,5) node [anchor=north west][inner sep=0.75pt]   [align=left] {\textbf{{\large VICON motion capture markers}}};

			\end{tikzpicture}
		}
		\caption{Quadrotor with Vicon motion capture markers used for indoor autonomous flight experiments~\cite{yu2024modular}.}
		\label{fig:UAV}
	\end{figure}
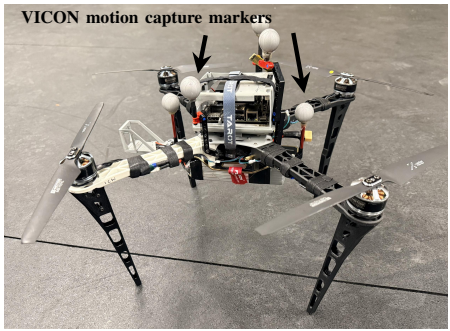
	
	\begin{figure*}[t]
		\centering
		\includegraphics[width=1 \linewidth]{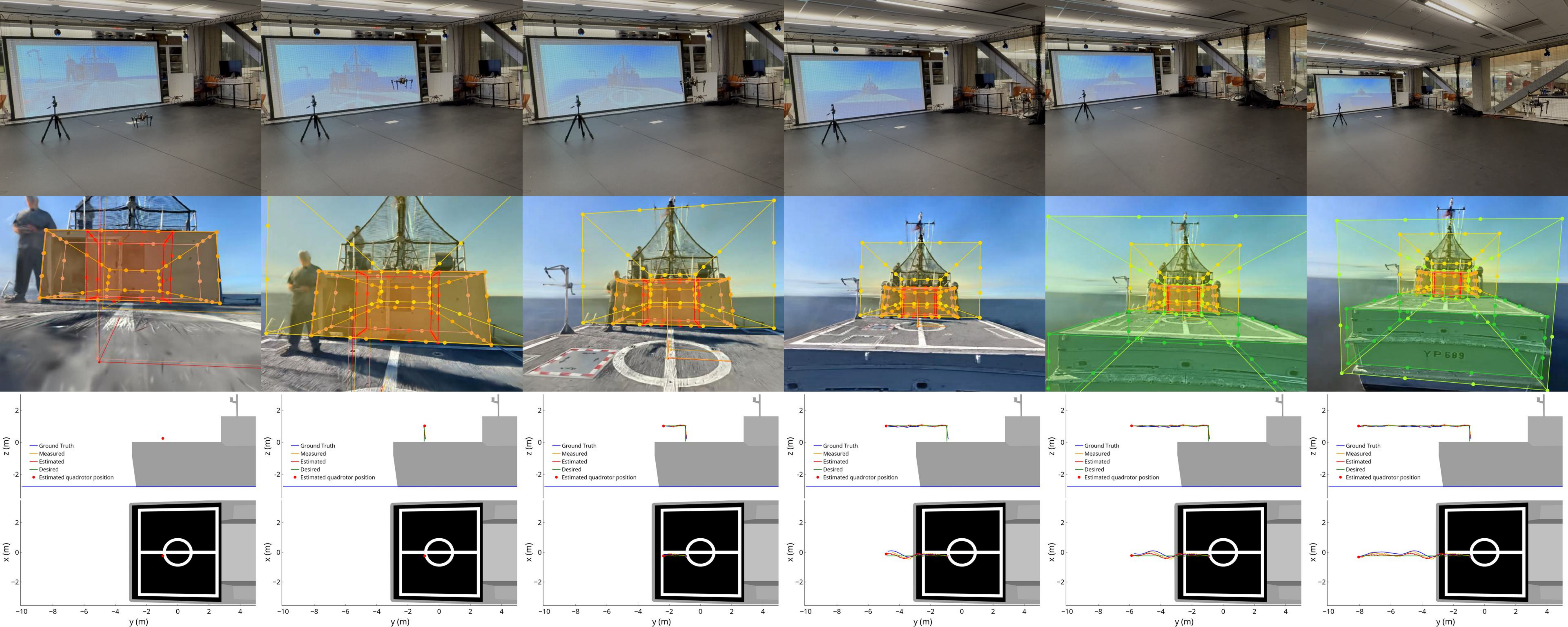}
		\caption{Vision-in-the-loop autonomous flight experiments: real-time photorealistic image generation, displayed on a projector screen for visualization.
			A video of the experiments can be found at
			\href{https://youtu.be/dfWmVE0wvoE}{https://youtu.be/dfWmVE0wvoE}.}
		\label{fig:UAVandTV}
	\end{figure*}

	The UAV platform (see \Cref{fig:UAV}) consists of a VectorNav VN100 IMU, T-Motor 700KV brushless motors, and MS1101 propellers, controlled by MikroKopter BL-Ctrl v2 ESCs for stable and precise actuation.
	A custom PCB manages power distribution, communication interfaces (I2C, UART), and sensor integration, powered by a 14.8V 6000mAh Li-Po battery.
	
	The flight software runs on the onboard Jetson and employs a multi-threaded architecture for real-time TNN inference, sensor processing, state estimation, and motor control.
	For the DKF implementation, the maximum supported measurement delay is set to $\tau_{\max}=\SI{5.5}{s}$, and the IMU buffer length is configured accordingly.
	The pose measurement frequency is \SI{5}{Hz}.
	The DKF fuses delayed pose estimates with 200~Hz IMU measurements to recover the state.
	Geometric control~\cite{LeeLeoPICDC10,GamLeeAJDSMC22} computes rotor thrust commands from this state estimate.
	The software (multi-threaded C++) handles estimation, control, and communication~\cite{KaniDelKal}.
	A GUI provides real-time monitoring, mission planning, and data visualization, with Wi-Fi enabling communication between the UAV and the ground station.

	\section{Autonomous Flight Experiments}\label{sec:Results}
	
	\subsection{Experimental Setup and Mission Profile}
	
	Autonomous flight experiments were conducted in an indoor motion-capture facility using the vision-in-the-loop pipeline.
	The mission consists of autonomous takeoff, tracking a straight 7.5~m out-and-back trajectory at a fixed altitude of 1.0~m (referenced in the virtual ship frame), and autonomous landing.
	A 3D visualization of the virtual ship environment and executed trajectory is shown in \Cref{fig:3Dship}, with corresponding top and side views provided in \Cref{fig:2Dship}.
	
	Performance is evaluated using two complementary error categories:
	\emph{estimation error}, defined as the difference between the DKF state estimate and Vicon ground truth, and
	\emph{control error}, defined as the difference between the desired trajectory and the measured state.
	Mean absolute error (MAE) is reported for position, velocity, and attitude, as summarized in \Cref{tab:eval}.
	
	\subsection{Experimental Results}
	
	\paragraph{Estimation Performance}
	
\begin{figure}[!t]
		\centering
		\includegraphics[width=1\linewidth]{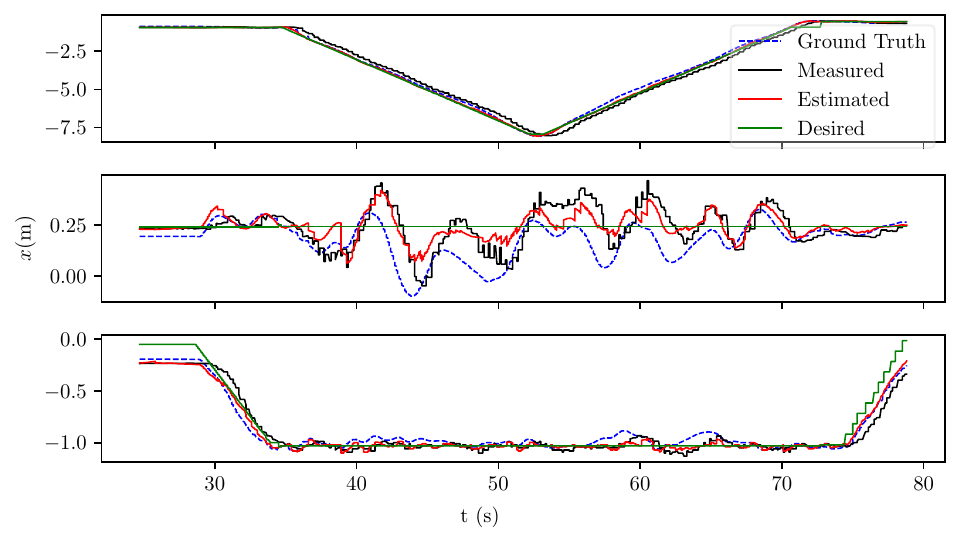}
		\caption{Position (autonomous flight test): the TNN-MO predicted position measurement (black) and the estimated position (red) are compared against the ground-truth Vicon position (blue) and the desired position (green).}
		\label{fig:pos}
	\end{figure}

\begin{figure}[!t]
		\centering
		\includegraphics[width=1\linewidth]{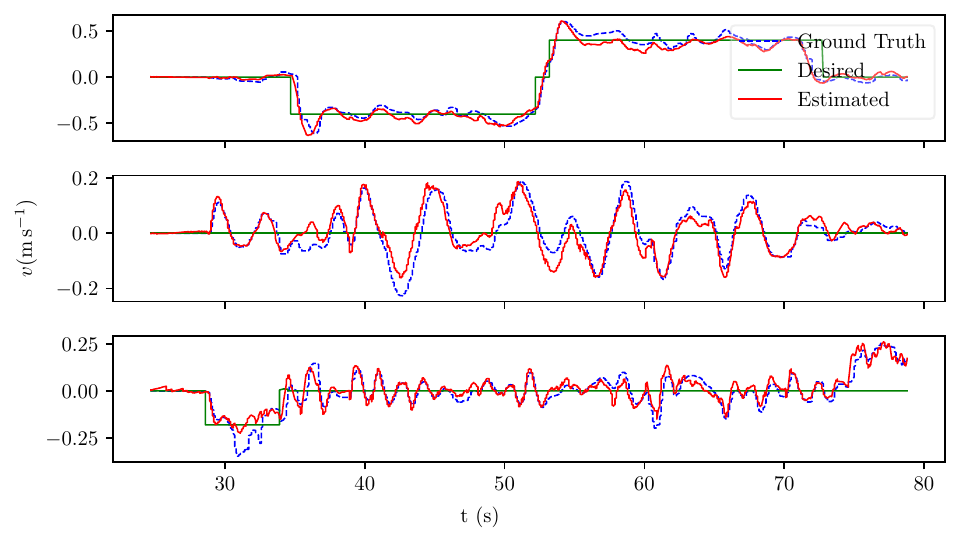}
		\caption{Velocity (autonomous flight test): the estimated velocity (red) is compared against the ground-truth Vicon velocity (blue) and the desired velocity (green).}
		\label{fig:vel}
	\end{figure}

\begin{figure}[!t]
		\centering
		\includegraphics[width=1\linewidth]{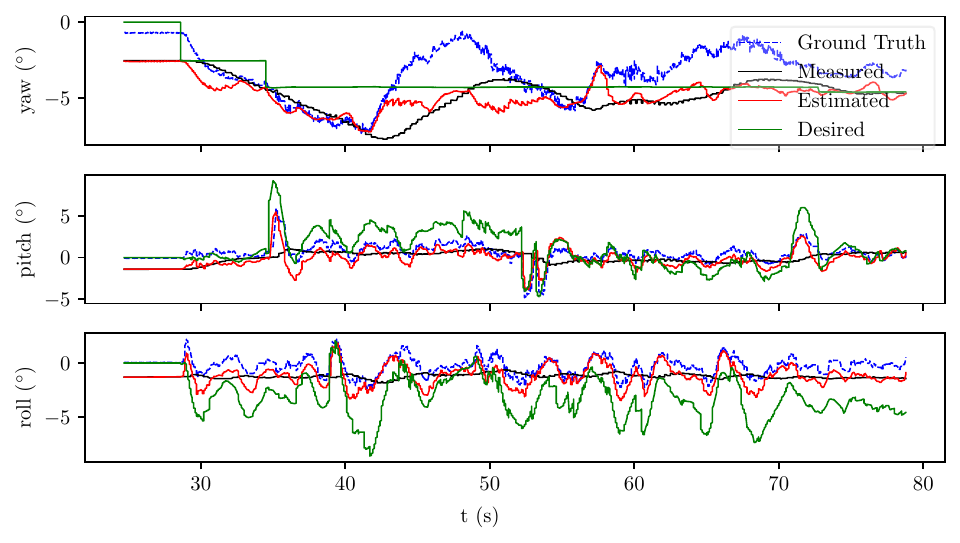}
		\caption{Attitude (autonomous flight test): the TNN-MO predicted attitude measurement (black) and the estimated attitude (red) are compared against the ground-truth Vicon attitude (blue) and the desired attitude (green).}
		\label{fig:atti}
	\end{figure}

	As summarized in \Cref{tab:eval} and illustrated in the position, velocity, and attitude time histories (\Cref{fig:pos,fig:vel,fig:atti}), the DKF achieves low estimation error despite delayed vision measurements.
	The position MAE is 0.066~m, velocity MAE is 0.032~m/s, and attitude MAE is 2.13$^\circ$.
	In \Cref{fig:pos}, the delayed TNN-MO measurements (black) and DKF estimates (red) are compared against Vicon ground truth (blue).
	\Cref{fig:vel} shows consistent velocity estimation, while \Cref{fig:atti} demonstrates accurate recovery of roll, pitch, and yaw.
	These results confirm that delayed monocular TNN-MO measurements are effectively fused with high-rate IMU data to recover accurate current-time state estimates.
	
	\paragraph{Closed-Loop Control Performance}
	
\begin{figure}[ht!] 
		\centering
		\includegraphics[width=1\linewidth]{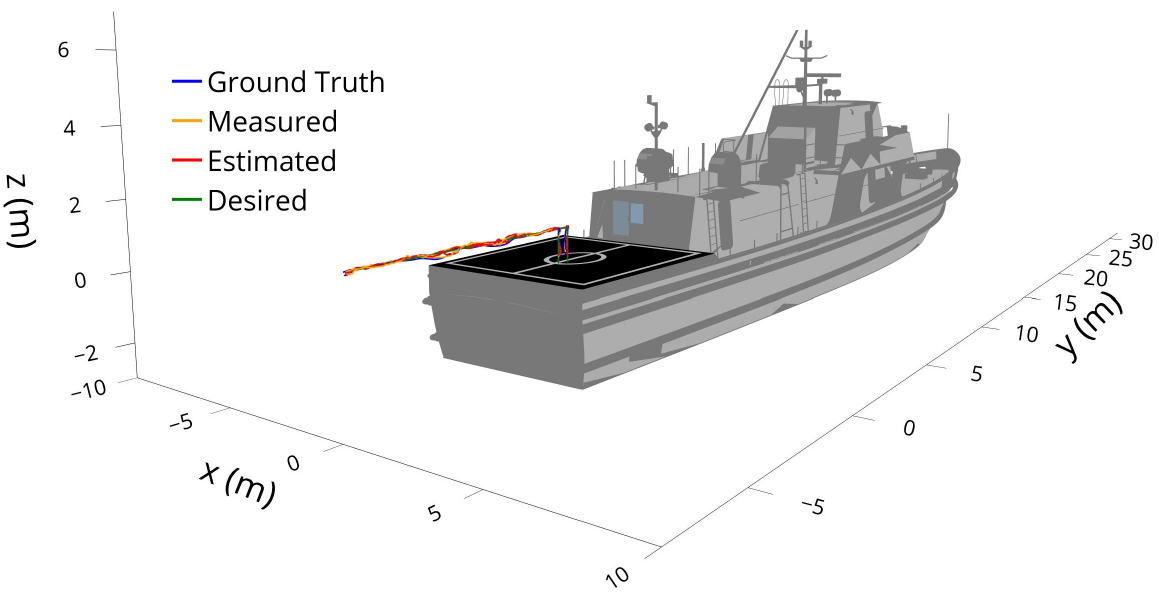}
		\caption{3D view of the ship for the autonomous flight test}
		\label{fig:3Dship}
	\end{figure}

\begin{figure}[ht!]
		\centering
		\includegraphics[width=0.9\linewidth]{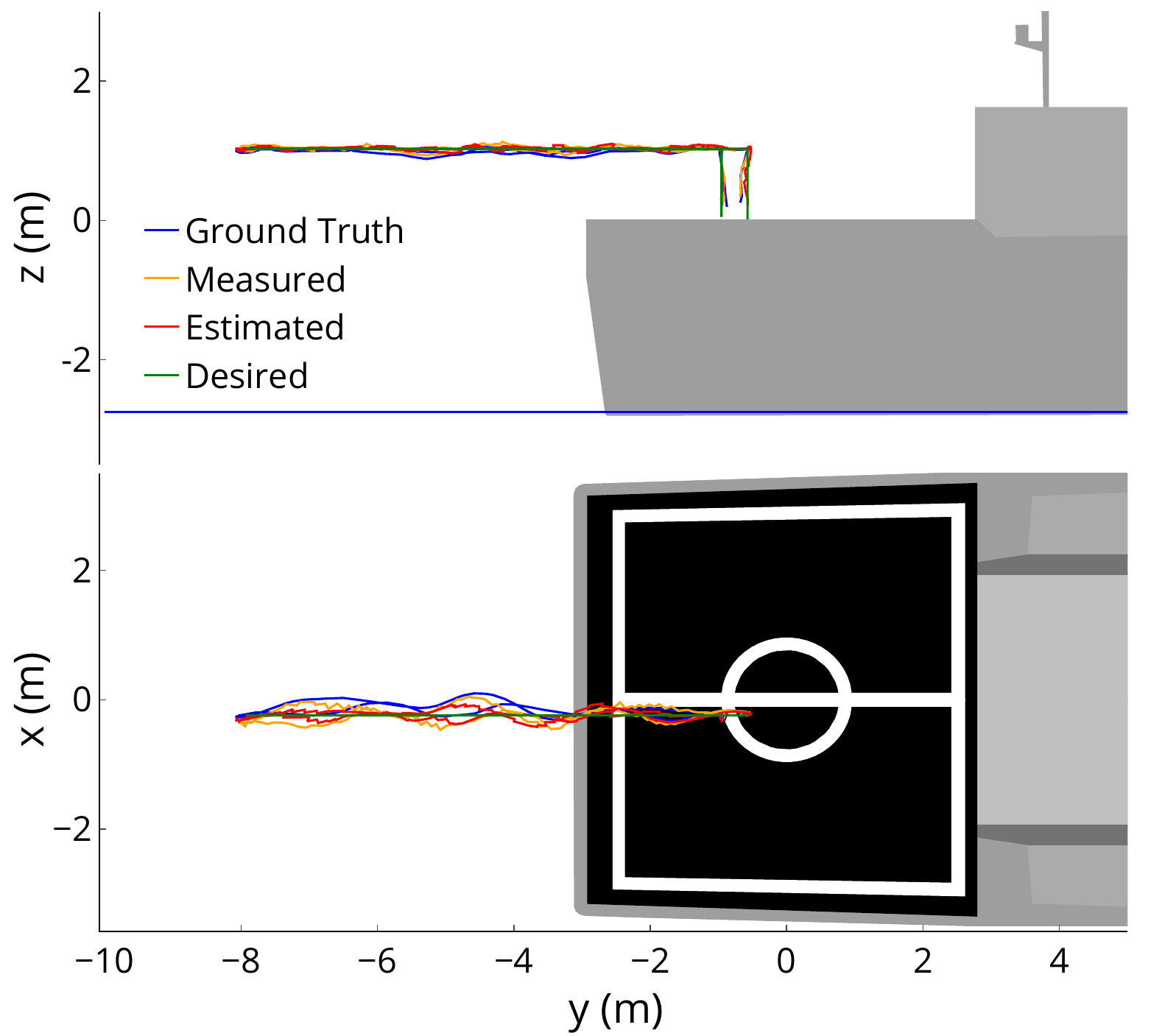}
		\caption{Top and side view of the ship for the autonomous flight test}
		\label{fig:2Dship}
	\end{figure}

	Closed-loop tracking performance remains stable throughout the mission.
	The control MAE is 0.089~m in position, 0.062~m/s in velocity, and 4.00$^\circ$ in attitude (see \Cref{tab:eval}).
	Trajectory tracking in the virtual ship frame is shown in \Cref{fig:3Dship,fig:2Dship}, where the UAV performs autonomous takeoff, tracks the 7.5~m out-and-back path while maintaining altitude, and lands reliably.
	The desired and measured trajectories are compared in \Cref{fig:pos,fig:vel,fig:atti}, demonstrating stable geometric control under delayed vision updates.
	These results confirm that the DKF state estimate is sufficiently accurate to support closed-loop autonomy.
	
	\paragraph{Robustness to Latency and Communication Variability}
	
	During experiments, Wi-Fi transmission latency fluctuated, introducing time-varying end-to-end delay in the vision pipeline.
	To ensure robustness, the maximum supported measurement delay was set to $\tau_{\max}=\SI{0.55}{s}$.
	A conservative fixed inference delay of 0.4~s was imposed, and two parallel TNN-MO instances were executed to maintain an effective 5~Hz vision update rate.
	Including transmission delay, the total latency was approximately 0.55~s under worst-case conditions.
	
	Despite variable communication delay and asynchronous measurements, the DKF maintained estimator consistency and enabled stable quadrotor flight, as evidenced by the smooth state and tracking responses in \Cref{fig:pos,fig:vel,fig:atti}.
	These results validate that delayed deep monocular vision measurements can reliably support real-time autonomous flight under realistic embedded and communication constraints.

	\begin{table}[t]
		\centering
		\caption{Evaluation Metrics}
		\label{tab:eval}
		\small
		\begin{tabularx}{\linewidth}{lXccc}
			\toprule
			Metric & Error & MSE & RMSE & MAE \\
			\midrule
			\multicolumn{5}{l}{\textbf{Estimation}}\\
			Position (m) & $\bar{x}-x$ & 0.0082 & 0.0907 & 0.0660 \\
			Velocity (m/s) & $\bar{v}-v$ & 0.002 & 0.043 & 0.032 \\
			\addlinespace[0.3em]
			Attitude (deg) & \begin{tabular}[c]{@{}l@{}}$\left\|\log\!\left(\bar{R}^\top R\right)^\vee\right\|$\end{tabular} & 5.30 & 2.30 & 2.13 \\
			\midrule
			\multicolumn{5}{l}{\textbf{Control}}\\
			Position (m) & $x_{\text{desired}}-x$ & 0.0152 & 0.1235 & 0.0894 \\
			Velocity (m/s) & $v_{\text{desired}}-v$ & 0.008 & 0.087 & 0.062 \\
			\addlinespace[0.3em]
			Attitude (deg) & \begin{tabular}[c]{@{}l@{}}$\left\|\log\!\left(R_{\text{desired}}^\top R\right)^\vee\right\|$\end{tabular} & 17.39 & 4.17 & 4.00 \\
			\bottomrule
		\end{tabularx}
	\end{table}
	
	\section{Conclusions}
	
	This work demonstrates fully autonomous indoor flight using a hardware-validated vision-in-the-loop framework for maritime UAVs.
	The proposed architecture integrates photorealistic 3D Gaussian Splatting (3DGS) rendering, onboard deep transformer-based pose estimation, and delayed Kalman filter (DKF) fusion to provide consistent current-time state estimates for geometric control.
	Through real-time embedded deployment and physical flight testing, we validate the complete perception–estimation–control loop under realistic communication latency and computational constraints.
	
	Beyond demonstrating feasibility, the experiments quantify the impact of end-to-end perception latency arising from rendering, communication, and edge inference.
	The results show that explicit delay compensation via the DKF is essential for maintaining estimator consistency and closed-loop stability when using delayed deep monocular pose estimates.
	This framework establishes a safe and hardware-realistic intermediate stage between simulation-only development and at-sea shipboard deployment.
	
	Future work will transition the system to real maritime environments, including autonomous flight experiments on a moving ship and evaluation under dynamic sea-state and airwake disturbances.
	
	\section*{Acknowledgement}
	
	This research was conducted in part on the U.S. Naval Academy’s research vessel YP689.
	The authors gratefully acknowledge the U.S. Naval Academy and the crew of YP689 for their support.
	We also thank the NAVAIR team for providing the opportunity and support for this research.

	\bibliography{ICUAS26}
	\bibliographystyle{IEEEtran}
	
\end{document}